\documentclass[10pt,twocolumn,letterpaper]{article}
\usepackage{cvpr}              
\usepackage{graphicx}
\usepackage{amsmath}
\usepackage{amssymb}
\usepackage{booktabs}
\usepackage{kotex}
\usepackage{bm}
\usepackage{environ}
\usepackage[pagebackref,breaklinks,colorlinks]{hyperref}

\newcommand{\round}[1]{\ensuremath{\lfloor#1\rceil}}

\usepackage{multirow}
\usepackage{booktabs}

\usepackage[capitalize]{cleveref}
\crefname{section}{Sec.}{Secs.}
\Crefname{section}{Section}{Sections}
\Crefname{table}{Table}{Tables}
\crefname{table}{Tab.}{Tabs.}

\def\cvprPaperID{8071} 
\def\confName{CVPR}
\def\confYear{2022}

\begin{document}
\title{Pooling Revisited: Your Receptive Field is Suboptimal}

\author{
Dong-Hwan Jang$^1$ \qquad Sanghyeok Chu$^1$ \qquad Joonhyuk Kim$^1$ \qquad Bohyung Han$^{1,2}$ \\
$^1$ECE \& $^1$ASRI \& $^{2}$IPAI, ~Seoul National University\\
 {\tt\small \{jh01120, sanghyeok.chu, kjh42551, bhhan\}@snu.ac.kr}
}

\maketitle


\begin{abstract}

The size and shape of the receptive field determine how the network aggregates local information and affect the overall performance of a model considerably. 
Many components in a neural network, such as kernel sizes and strides for convolution and pooling operations, influence the configuration of a receptive field. 
However, they still rely on hyperparameters, and the receptive fields of existing models result in suboptimal shapes and sizes.
Hence, we propose a simple yet effective Dynamically Optimized Pooling operation, referred to as DynOPool, which optimizes the scale factors of feature maps end-to-end by learning the desirable size and shape of its receptive field in each layer.
Any kind of resizing modules in a deep neural network can be replaced by the operations with DynOPool at a minimal cost.
Also, DynOPool controls the complexity of a model by introducing an additional loss term that constrains computational cost.
Our experiments show that the models equipped with the proposed learnable resizing module outperform the baseline networks on multiple datasets in image classification and semantic segmentation.%
\end{abstract}


\section{Introduction}
\label{sec:intro}

Despite the unprecedented success of deep neural networks in various applications including computer vision~\cite{simonyan2015very, he2016deep, sandler2018mobilenetv2, liu2021swin}, natural language processing~\cite{devlin2019bert, radford2019language}, robotics~\cite{labbe2021single}, and bioinformatics~\cite{ji2021learning}, the design of the optimal network architecture is still a challenging problem. 
While several handcrafted models exhibit impressive performance in various domains, there have been substantial efforts to identify the optimal neural network architecture with associated operations automatically~\cite{kim2019fine, liu2018darts, tan2019efficientnet, kang2020operation}.
However, hand-engineered architectures are prone to be suboptimal and suffer from weak generalizability while the approaches based on neural architecture search either incur a huge amount of training cost or achieve minor improvement due to limited search space.

Researchers have been investigating powerful and efficient operations applicable to deep neural networks, which include convolutions, normalizations, and activation functions.
However, they have not paid much attention to pooling operations despite their simplicity and effectiveness in aggregating local information.
Also, although some operations have been adopted to design receptive fields, which are critical to the overall performance of a network, the optimization of their sizes and shapes is not studied thoroughly.
The receptive field is determined by several factors in deep neural networks such as the depth of a model, strides of operations, types of convolutions, etc.
As the methods to adjust the receptive field of an operation, variants of convolution operations~\cite{noh2015learning, YuKoltun2016, dai2017deformable} or special architectures with multi-resolution branches~\cite{he2015spatial, YuanCW19} are widely adopted.
However, these approaches rely on delicately human-engineered hyperparameters or time-consuming neural architecture search~\cite{zoph2016neural, zoph2018learning}.

To alleviate the suboptimality of human-engineered architectures and operations, we propose Dynamically Optimized Pooling operation (DynOPool), which is a learnable resizing module that replaces standard resizing operations.
The proposed module finds the optimal scale factor of a receptive field for the operations learned on a dataset, and, consequently, resizes the intermediate feature maps in a network to proper sizes and shapes.
This relieves us from the delicate design of hyperparameters such as stride of convolution filters and pooling operators.

Our contributions are summarized as follows:
\begin{itemize}
\item Our work tackles the limitations of existing scaling operators in deep neural networks that depend on predetermined hyperparameters. We point out the importance of finding the optimal spatial resolutions and receptive fields in intermediate feature maps, which are still under-explored in designing neural architectures.
\item We propose DynOPool, a learnable resizing module that finds the optimal scale factors and receptive fields of intermediate feature maps.
DynOPool identifies the best resolution and receptive field of a certain layer using a learned scaling factor and propagates the information to the subsequent ones leading to scale optimization across the entire network. 
\item We demonstrate that the model with DynOPool outperforms the baseline algorithms on multiple datasets and network architectures in the image classification and semantic segmentation tasks. 
It also exhibits desirable trade-offs between accuracy and computational cost.
\end{itemize}

Our paper is organized as follows. 
Section~\ref{sec:related} presents existing related works and Section~\ref{sec:motivation} introduces our motivation for optimizing the size and shape of the receptive field and feature map.
We describe the technical details of DynOPool in Section~\ref{sec:method} and experimental results in Section~\ref{sec:experiments}. 
Last, we conclude this work and discuss future works in Section~\ref{sec:conclusion}.


\section{Related Works}
\label{sec:related}
\paragraph{Neural architecture search}
Neural Architecture Search (NAS)~\cite{zoph2016neural, zoph2018learning, pham2018efficient, liu2018darts, luo2018neural} is an AutoML method that optimizes the structure of a deep neural network architecture by formulating a hyperparameter setting with human inductive bias as a learnable procedure.
Previous approaches based on reinforcement learning~\cite{zoph2016neural, zoph2018learning, pham2018efficient} require huge amount of GPU time.
Although several methods have been proposed to accelerate the search process by sharing weights~\cite{pham2018efficient} or gradient-based optimization~\cite{liu2018darts, luo2018neural}, they are still suboptimal due to search space constraints.
There exist a couple of prior works to search for input resolutions~\cite{huang2019gpipe,tan2019efficientnet}, but finding the optimal feature size and shape for each layer is still a challenging problem.
\vspace{-0.2cm}

\paragraph{Dynamic kernel shape}
Recent approaches~\cite{jeon2017active, papandreou2015modeling, pintea2021resolution,romero2021ckconv,romero2022flexconv} adopt variants of convolutions that learn the sizes of receptive fields dynamically.
N-Jet~\cite{pintea2021resolution} employs Gaussian derivative filters to adapt kernel size using the scale-space theory.
CKConv~\cite{romero2021ckconv} uses a continuous kernel parameterization trick to implement kernels of diverse sizes without additional cost.
Similarly, FlexConv~\cite{romero2022flexconv} utilizes the implicit neural representation to generate large-bandwidth filters of varying sizes.
These methods identify the optimized receptive fields by learning filter sizes while our approach does it via learning the size of the feature map.
\vspace{-0.2cm}

\paragraph{Learnable resizing modules}
Shape Adaptor~\cite{liu2020shape} controls the receptive field by direct learning of the feature map size.
It proposes a differentiable resizing module applicable to a linear combination of a pooled feature map with a ratio (\eg 0.5 or 1.5) and a non-pooled map.
However, the resizing module is limited to selecting one of the pre-defined ratios for upsampling or downsampling, and processing the symmetric resizing only.
Recently, DiffStride~\cite{riad2022learning} presents a spectral pooling method to determine the optimal stride of the pooling layer.
They find an appropriate feature map size and shape by replacing downsampling in the spatial domain with cropping in the frequency domain, where the cropping window size is optimized.


%
\begin{figure*}[t]
	\centering
	\footnotesize
	\setlength{\tabcolsep}{1pt}
	\scalebox{1.05}{
	\begin{tabular}{ccc}		
		\includegraphics[height=3.1cm]{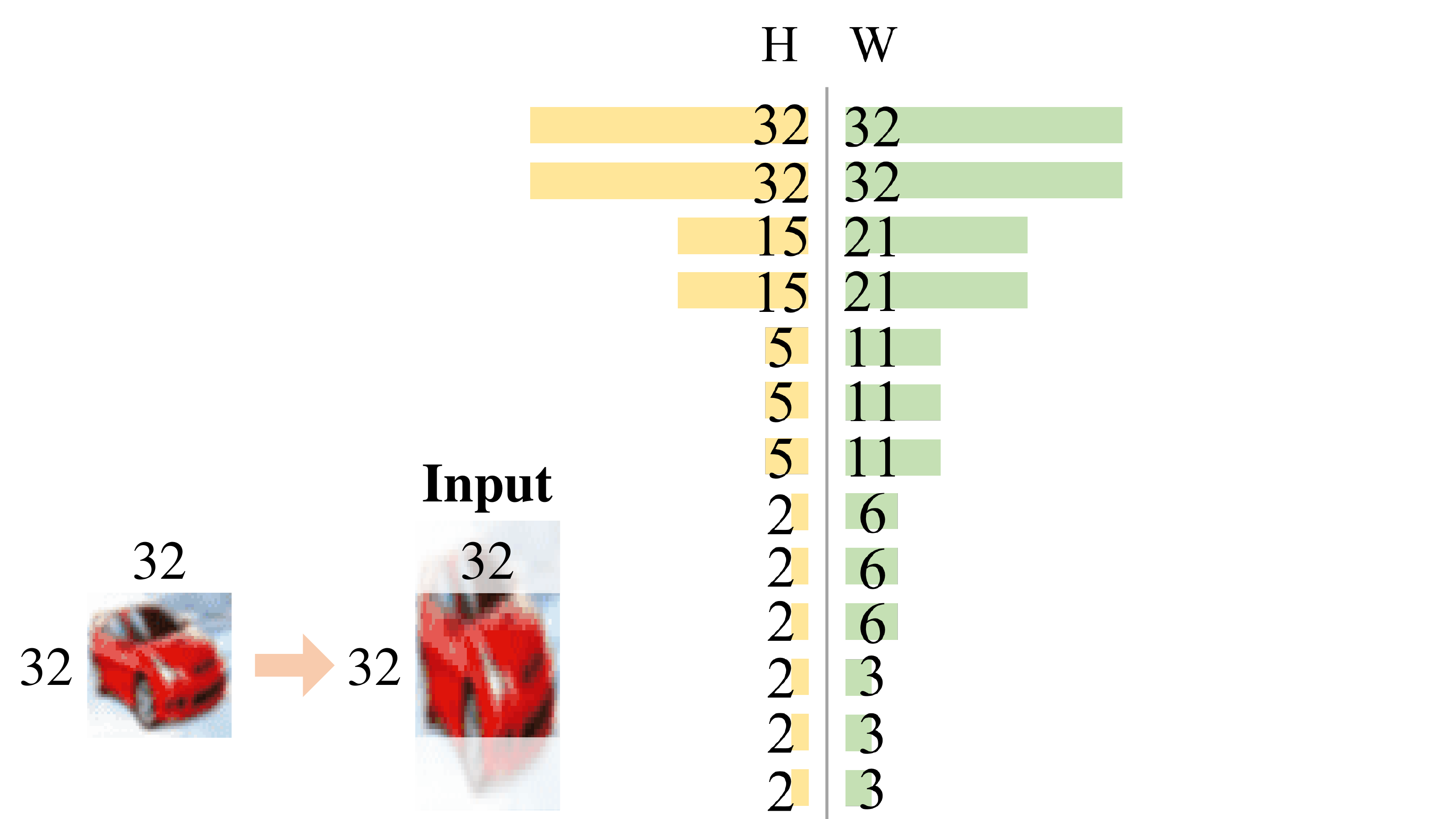} &
		\includegraphics[height=3.1cm]{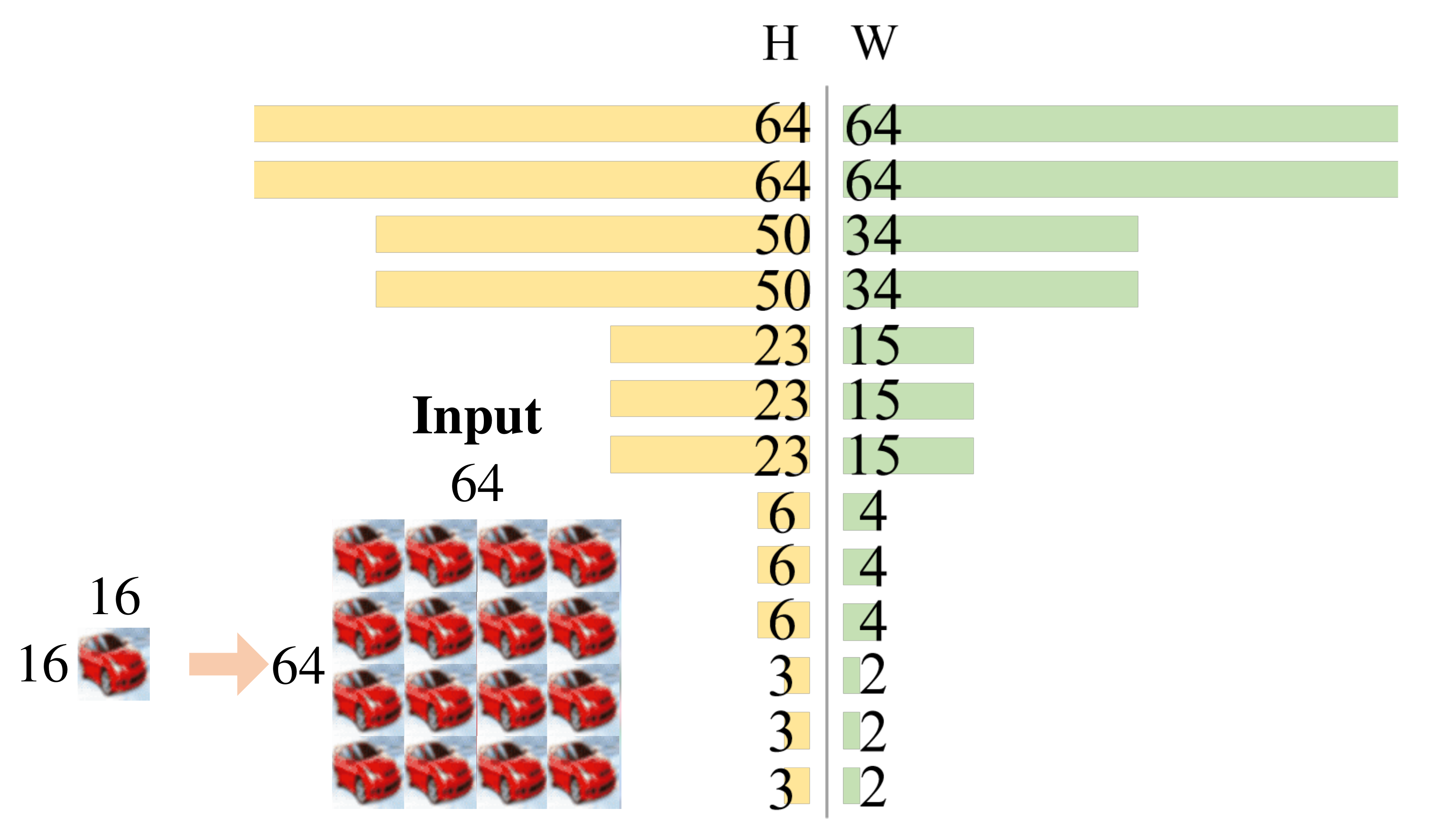} &
		\includegraphics[height=3.1cm]{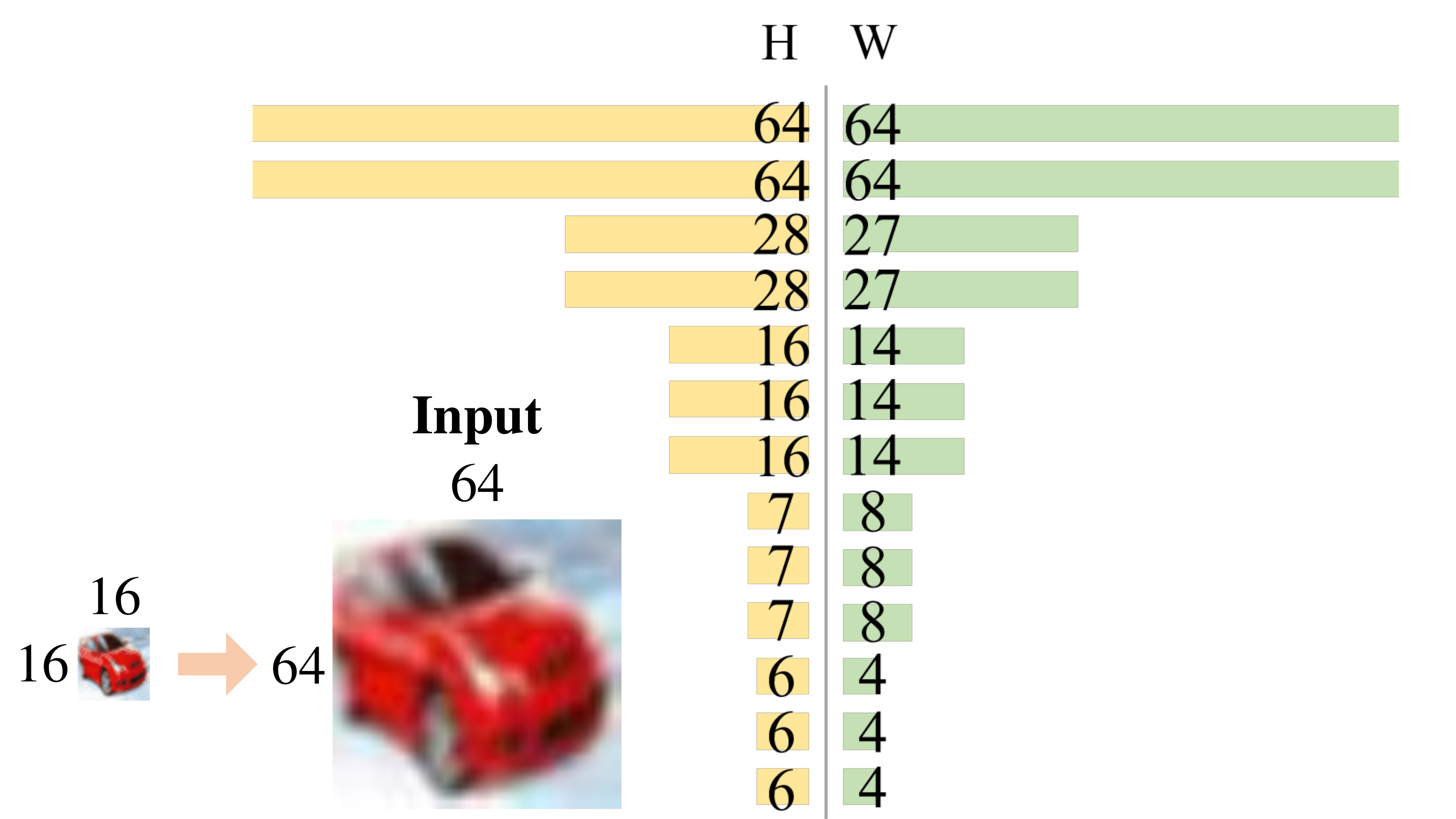} \vspace{0.2cm} \\
 		\shortstack{\textbf{0.30G / 71.18\%} (0.31G / 69.49\%) \\ (a) CIFAR-stretch-V } 
           & \shortstack{\textbf{1.24G / 60.06\%} (1.26G / 56.70\%) \\ (b) CIFAR-tile}  
           & \shortstack{\textbf{1.18G / 67.82\%} (1.26G / 65.49\%) \\ (c) CIFAR-large}  
		\end{tabular}
		}
    \caption{We conduct toy experiments with three different synthetic datasets made from CIFAR-100:
    (a) randomly crop a vertically stretched image (b) tile a downsized image in a 4$\times$4 grid (c) enlarge a downsized image. Although the contents are almost the same, the optimal size and shape of each feature map is greatly different depending on the characteristics of input images. Unlike the human-designed model, which has fixed feature map sizes, our models adjust the feature map sizes to maintain the optimal amount of information in each feature map, leading to improved performance. The numbers in bold face fonts are GMACs and the accuracy given by DynOPool while the numbers in parentheses are from human-designed models.}
	\label{fig:proof_of_concepts}	
\end{figure*}

\section{Motivation}
\label{sec:motivation}

The information in an image is spread over various levels of locality, and the CNN learns patterns with diverse scales using a series of kernels to learn strong representations. 
Since the sizes and shapes of semantically meaningful patterns differ greatly for each image, it is important to identify proper receptive fields and extract useful information using the receptive fields from an image. However, the exploration of the optimal receptive field has not been studied actively and the use of adaptive feature map size has hardly been discussed so far although several previous works indirectly learn the receptive field size via other methods such as neural architecture search or design learnable receptive fields with excessive constraints.
This section presents why a conventional receptive field with a fixed size and shape is suboptimal and discusses how DynOPool tackles this issue through toy experiments with VGG-16~\cite{simonyan2015very} on CIFAR-100~\cite{krizhevsky2009learning}.

\subsection{Asymmetrically Distributed Information}

The optimal receptive field shape changes according to the spatial information asymmetry inherent in the datasets.
For example, barcode images have no information along the vertical direction because the same value is repeated in the direction.
Therefore, it would be desirable to concentrate on the horizontal direction to represent the barcode images better.
The problem is that, except for the images with the prior information like barcodes, the inherent asymmetry is not measurable in most cases.
Also, input resizing, often used as preprocessing, sometimes leads to information asymmetry.
In human-designed networks, the aspect ratio of an image is often adjusted to satisfy the input specifications of models.
However, the receptive fields in such networks are not designed to handle the operations.

To address the potential of the proposed approach, DynOPool, we perform experiments on CIFAR-stretch-V, a toy dataset in which images of CIFAR-100 are vertically stretched twice in the vertical direction and cropped randomly to a size 32x32.
As shown in Figure~\ref{fig:proof_of_concepts}(a), wide feature maps whose shapes are dynamically optimized by DynOPool, achieve improved performance by extracting valuable information more in the horizontal direction, compared to the human-designed model.

\subsection{Densely or Sparsely Distributed Information}
The level of locality is another interesting component for designing optimal models. 
CNNs learn the complex representations from an image by aggregating local information in a cascaded manner.
However, the importance of the local information depends heavily on the properties of each image.
For example, when an image is blurred, most of the meaningful micro patterns, such as the texture of an object, are wiped out.
In this case, it would be better to extend the receptive field in early layers and concentrate on global information.
On the other hand, if an image contains plenty of class-specific information in local details, \textit{e.g.}, texture, recognizing local patterns would be more important.

To verify the hypothesis, we construct two variants of the CIFAR-100 dataset, CIFAR-tile and CIFAR-large, as shown in Figure~\ref{fig:proof_of_concepts}.
To this end, we first downsample the original images in CIFAR in half and construct $16 \times 16$ images.
Then, we tile the downsampled image in a $4 \times 4$ for CIFAR-tile, and upsample the downsampled images to size $64 \times 64$ for CIFAR-large.

As illustrated in Figure~\ref{fig:proof_of_concepts}(b) and (c), our models outperform the human-designed model by large margins.
Although both datasets are constructed with the same base images of size $16 \times 16$, the learned networks by DynOPool have notably different shapes; our model trained on CIFAR-tile has larger feature maps than the model trained on CIFAR-large in the early layers.
Note that DynOPool for the CIFAR-tile prefers to employ small receptive fields at the beginning of the network because the tiled objects are very small.
On the other hand, our model for the CIFAR-large is encouraged to have large receptive fields in the low level because the input image is magnified from a small one and it makes sense to observe large areas in the early layers.


\section{Proposed Method}
\label{sec:method}

We discuss the proposed learnable resizing module, referred to as DynOPool, in detail, which includes its concept, optimization, and practical benefits.

\subsection{Dynamically Optimized Pooling (DynOPool)}
\label{sec:dynapool}
The resizing module in DynOPool, which accepts an input feature map, $x_\text{in} \in \mathbb{R}^{H_\text{in} \times W_\text{in}}$, and returns a resized output, $x_\text{out} \in \mathbb{R}^{H_\text{out} \times W_\text{out}}$, is defined and optimized as follows.

\subsubsection{Design of DynOPool}

\begin{figure}[t]
    \centering
    \includegraphics[width=0.475\textwidth]{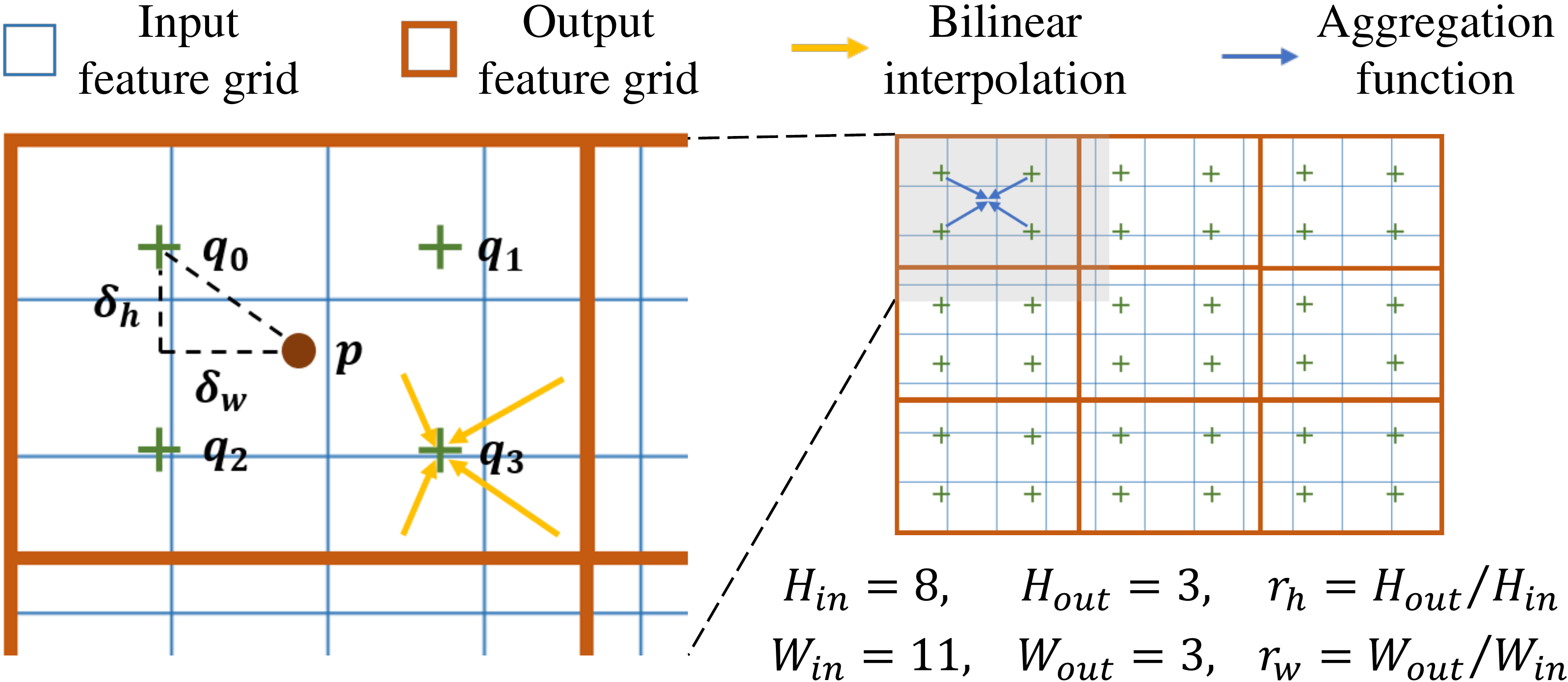}
    \caption{Overview of the proposed resizing module, DynOPool (best viewed in color). We optimize the scale factor $\bm{r}= (r_{h}, r_{w})$ between a pair of input and output feature maps, denoted by $x_\text{in}$ and $x_\text{out}$, respectively. The brown dot $\bm{p}$ represents the center of a grid cell in $x_\text{out}$ while the green crosses indicate four query points $\bm{q}$ in the same cell. The representation of $q_{i}$ is given by bilinear interpolation of the features corresponding to the four nearest pixels in $x_\text{in}$. An output feature of a grid cell in $x_\text{out}$ is derived by the feature aggregation of the four query points, where a simple aggregation function such as max-pooling is typically employed.}
    \label{fig:dynopool}
\end{figure}

Figure~\ref{fig:dynopool} illustrates how DynOPool works.
DynOPool first divides the feature map $x_\text{in}$ into an $H_\text{out} \times W_\text{out}$ grid as 
\begin{equation}
\begin{split}
   H_\text{out} &= \round{H_\text{in} \cdot r_{h}} \\
   W_\text{out} &= \round{W_\text{in} \cdot r_{w}},
\end{split}
\label{eq:scale}
\end{equation}
where $\bm{r} = (r_{h},r_{w})$ indicates the scale factor for height and width of a feature map and $\round{\cdot}$ is a round operation.
Assuming that $(-1,-1)$ and $(1,1)$ are the normalized coordinates of the top-left and bottom-right corners of $x_\text{in}$, the size of a grid cell in the output feature map becomes $\frac{2}{H_\text{out}} \times \frac{2}{W_\text{out}}$.%

Then, given a grid cell centered at $\bm{p}=(p_{h}, p_{w})$, the positions of the four query points are defined as
\begin{equation}
\begin{split}
    \bm{q} & = \left( p_{h} \pm \delta_{h}, \ p_{w} \pm \delta_{w} \right) \\
                & = \left( p_{h} \pm \frac{1}{4} \cdot \frac{2}{H_\text{out}}, \ p_{w} \pm  \frac{1}{4} \cdot \frac{2}{W_\text{out}} \right),
    \label{eq:delta}
\end{split}
\end{equation}
where $\bm{\delta} = (\delta_{h}, \delta_{w})$ denotes the displacement from $\bm{p}$.
The representation of each query point is given by bilinear interpolation of four nearest grid cells in $x_\text{in}$.
Then, DynOPool aggregates the four feature vectors and returns the output representation of each grid cell in $x_\text{out}$.
We choose max-pooling as an aggregation function, but any other function can replace max-pooling as long as it is effective to compute abstract representations from multiple local features.

The primary benefits of DynOPool with the optimized scale factor $\bm{r}$ are twofold.
First, the location of four query points $\bm{q}$ are also optimized because $\bm{\delta}$ is a function of $\bm{r}$.
Second, by obtaining the best resolution of an intermediate feature map through the optimization of $\bm{r}$, DynOPool adaptively controls the size and shape of receptive fields in deeper layers with other operators intact.
\vspace{-0.2cm}

\subsubsection{Optimization}

The rescaling module is defined by a combination of \eqref{eq:scale} and \eqref{eq:delta}, which are based on simple operations. 
However, the rounding operations are not differentiable and hinder the optimization procedure of DynOPool.
To remedy this issue, we leverage a differentiable quantization trick, which is a well-known continuous relaxation technique for discrete random variables~\cite{jang2016categorical, maddison2017concrete}.
Then the rescaling modules are given by reformulating the round functions as follows:
\begin{align}
    H_{\text{out}} &= \round{H_{\text{in}}\cdot r_h}+H_{\text{in}} \cdot r_h -sg(H_{\text{in}} \cdot r_h),  \label{eq:diffH} \\  
    W_{\text{out}} &= \round{W_{\text{in}}\cdot r_w}+W_{\text{in}} \cdot r_w -sg(W_{\text{in}} \cdot r_w),   \label{eq:diffW} 
\end{align}
where $sg(\cdot)$ indicates a stop gradient operator~\cite{bengio2013estimating}.
Note that \eqref{eq:diffH} and \eqref{eq:diffW} allow us to feedforward the original discrete values $\round{H_{\text{in}}\cdot r_h}$ and $\round{W_{\text{in}}\cdot r_w}$ while backpropagating through their continuous surrogate functions $H_\text{in} \cdot r_{h}$ and $W_\text{in} \cdot r_{w}$.

\begin{figure}[t]
    \centering
    \includegraphics[width=\linewidth]{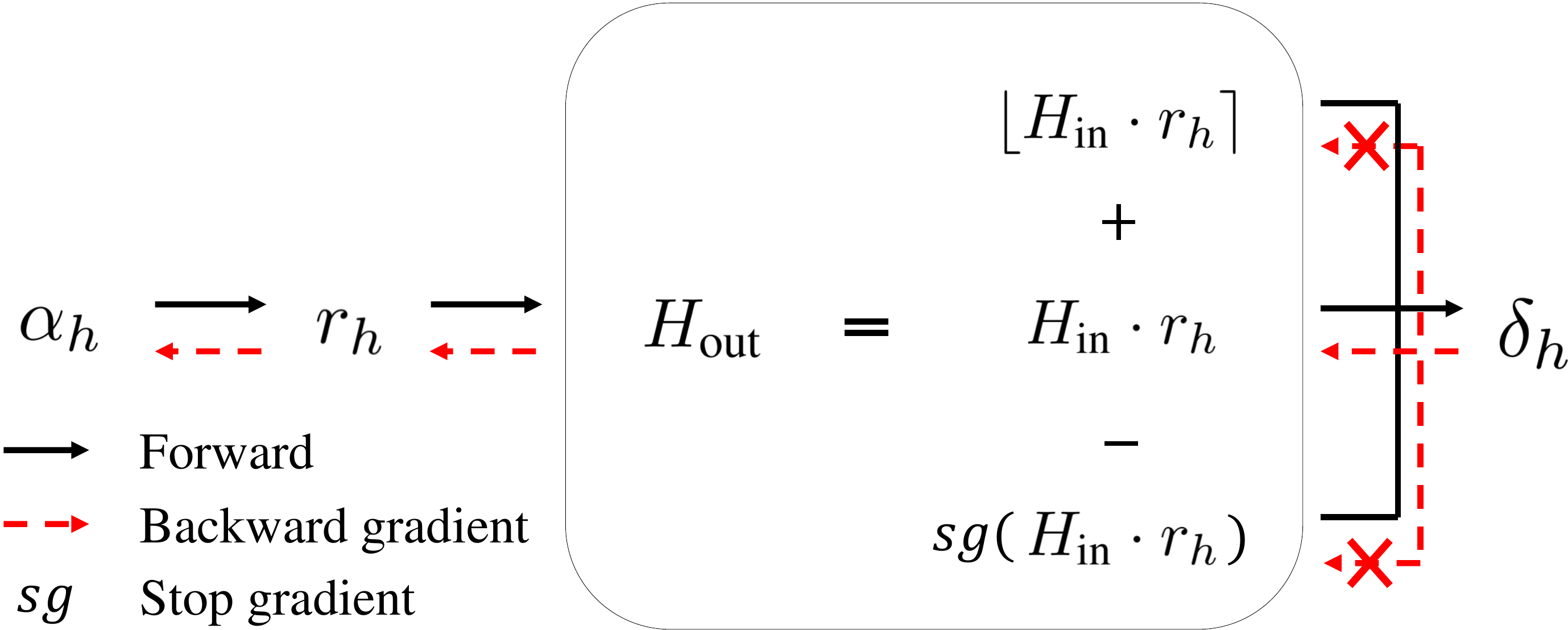}
    \caption{Computational flows inside DynOPool. Although the forward pass employs the discretized value, $\round{H_{\text{in}}\cdot r_h}$, its continuous counterpart ($H_{\text{in}}\cdot r_h$) is adopted in the backward pass to backpropagate the gradients into $\bm{\alpha}$. The same optimization process is applied with respect to width.}
    \label{fig:optimization}
\end{figure}

Although the optimization is now feasible, there remains an additional challenge in learning the scale factor $\bm{r}$.
As expressed in \eqref{eq:delta}, the rescaling module involves the displacement function, $\bm{\delta$}, which depends on $\bm{r}$.
However, the gradient with respect to $\bm{r}$ is unstable when either $r_h$ or $r_w$ is small because the gradient is inversely proportional to $r_{h}^{2}$ or $r_{w}^{2}$ as
\begin{equation}
\frac{d\delta_h}{dr_{h}} \propto -\frac{1}{r_{h}^{2}} ~~~\text{and}~~~ \frac{d\delta_w}{dr_{w}}  \propto -\frac{1}{r_{w}^{2}}
    \label{eq:gradient_r}
\end{equation}
Since this gradient explosion results in significant changes in the resolution of $x_\text{out}$ during training, we reparameterize $\bm{r}$ using $\bm{\alpha}=[\alpha_{h}, \alpha_{w}]$ as follows:
\begin{equation}
    [\alpha_{h}, \alpha_{w}] = [r_{h}^{-1}, r_{w}^{-1}].
    \label{eq:relationship_alpha_r}
\end{equation}
By defining $\bm{\alpha}$ as a learnable scale parameter and optimizing it instead of $\bm{r}$,  the training procedure is greatly stabilized in practice.
Figure~\ref{fig:optimization} illustrates the overall optimization process.


\subsection{Constraints for Model Complexity}
\label{sec:gmacsloss}
To maximize the accuracy of models, DynOPool sometimes has a large scale factor and increases the resolution of intermediate feature maps.
Therefore, to constrain computational cost and reduce model size, we introduce an additional loss term $\mathcal{L}_{\text{GMACs}}$, which is given by a simple weighted sum of layerwise GMACs counts at each training iteration $t$ as follows:
\begin{equation}
\begin{split}
   \mathcal{L}_{\text{GMACs}} & = \sum_{\ell=1}^{N} w_{\ell}^{t} \cdot \text{GMACs}[\ell] \\
                                  & = \sum_{\ell=1}^{N} \frac{H_\text{out}^{t \ (\ell)} \cdot W_\text{out}^{t \ (\ell)}}{H_{\text{out}}^{\text{0} \ (\ell)} \cdot W_{\text{out}}^{\text{0} \ (\ell)}} \cdot \text{GMACs}[\ell],
    \label{eq:gmacsloss_tmp}
\end{split}
\end{equation}
where $N$ is the total number of layers in the model, $\text{GMACs}[\ell]$ represents the GMACs counts in the $\ell^\text{th}$ layer at the initial state, and $w_{\ell}^{t}$ is the ratio between the feature map sizes in the $\ell^\text{th}$ layer at the initial stage ($H_{out}^{0 \ (\ell)}$, $W_{out}^{0 \ (\ell)}$) and the current training iteration $t$ ($H_{out}^{t \ (\ell)}$, $W_{out}^{t \ (\ell)}$). 
By definition, $\mathcal{L}_{\text{GMACs}}$ reflects the degree of the computational cost increase as the scale factor $\bm{r}$ changes during training, compared to the initial state of the model.

\subsection{Loss}
\label{sec:loss}

We train the model with DynOPool by a linear combination of the task-specific objectives ($\mathcal{L}_{\text{task}}$) and the proposed GMACs loss ($\mathcal{L}_{\text{GMACs}}$) as follows:
\begin{equation}
   \mathcal{L}_{\text{total}} = \mathcal{L}_{\text{task}} + \lambda \cdot \mathcal{L}_{\text{GMACs}},
    \label{eq:loss_total}
\end{equation}
where $\lambda$ is a hyperparameter that controls the computational complexity of a model and maintains the balance with the task-specific loss. 
The model is trained to maximize its performance by jointly learning the optimal spatial resolutions of intermediate feature maps. 


\begin{table*}[t!]
    \centering
    \small
    \caption{
    Top-1 accuracy (\%) and GMACs comparisons between human-designed models and models with DynOPool.
  The sizes and shapes of the feature maps for each block in the network architecture are also reported.
  DynOPool-S outperforms human-designed models with comparable GMACs in almost all cases. 
  Notably, DynOPool-S compresses the model up to 33\% lighter than the human-designed VGG-16 for the ImageNet dataset while maintaining the accuracy of the model.
  DynOPool-B outperforms the human-designed models with significant margins in all cases.
    }
    
    \label{tab:main}
    \scalebox{0.78}{
    \hspace{-2mm}
    \setlength\tabcolsep{1.5pt} 
    \begin{tabular}{clcclcclccl}
    \hline
    
    \toprule
    
    \multicolumn{2}{c}{\multirow{2}{*}{\textbf{Dataset}}}
    & \multicolumn{3}{c}{\textbf{FGVC-Aircraft}} 
    & \multicolumn{3}{c}{\textbf{CIFAR-100}} 
    & \multicolumn{3}{c}{\textbf{ImageNet}} 
    \\ \cline{3-11}
    \\ [-0.9em]
            
    &  
    & {Acc.} & {GMACs} & \hspace{16mm}{Feature map sizes} 
    & {Acc.} & {GMACs} & \hspace{5mm}{Feature map sizes}    
    & {Acc.} & {GMACs} & \hspace{16mm}{Feature map sizes} 
    \\ \cmidrule{1-11}\morecmidrules\cmidrule{1-11}
    
    \multicolumn{1}{c}{\multirow{4}{*}{\rotatebox{90}{VGG-16}}~~~}
    & Human
    & 85.3            & 15.40        
    & \scriptsize[224,224] [112,112] [56,56] [28,28] [14,14]   
    & 75.4            & 0.31        
    & \scriptsize[32,32] [16,16] [8,8] [4,4] [2,2]
    & 73.9            & 15.39 
    & \scriptsize[224,224] [112,112] [56,56] [28,28] [14,14]
    \\ \cline{2-11}
    \\ [-0.9em]
    & DynOPool-S
    & 87.0           & 13.90        
    & \scriptsize[224,224] [114,142] [52,53] [30,19] [17,7]    
    & 75.5           & 0.36 
    & \scriptsize[32,32] [21,14] [10,7] [5,4] [2,2]
    & 73.8           & 10.16 
    & \scriptsize[224,224] [88,87] [40,37] [24,23] [12,12]     
    \\ 
    \\ [-0.9em]    
    & DynOPool-B
    & \bf{87.4}            & 32.39        
    & \scriptsize[224,224] [127,256] [76,102] [46,37] [20,11]  
    & \bf{79.8}            & 1.71        
    & \scriptsize[32,32] [37,32] [21,18] [12,9] [7,4]
    & \bf{74.1}          & 20.92 
    & \scriptsize[224,224] [151,152] [67,68] [32,30] [15,13]       
    \\ \cmidrule{1-11}\morecmidrules\cmidrule{1-11}
    
    \multicolumn{1}{c}{\multirow{4}{*}{\rotatebox{90}{ResNet-50}}~~}
    & Human      
    & 81.6            & 4.12        
    & \scriptsize[224,224] [56,56] [28,28] [14,14] [7,7]   
    & 78.5            & 1.31        
    & \scriptsize[32,32] [16,16] [8,8] [4,4]
    & 77.2           & 4.11 
    & \scriptsize[224,224] [56,56] [28,28] [14,14] [7,7]     
    \\ \cline{2-11} 
    \\ [-0.9em]
    & DynOPool-S
    & 82.3            & 3.57        
    & \scriptsize[224,224] [58,63] [18,17] [9,4] [4,2] 
    & 80.3            & 1.01        
    & \scriptsize[32,32] [10,9] [5,4] [2,2]
    & 77.6           & 6.20 
    & \scriptsize[224,224] [71,71] [27,26] [12,11] [4,2]          
    \\
    \\ [-0.9em]
    & DynOPool-B
    & \bf{87.2}            & 38.53        
    & \scriptsize[224,224] [225,210] [68,66] [16,17] [4,4]  
    & \bf{80.6}            & 1.73        
    & \scriptsize[32,32] [18,17] [7,6] [2,3]
    & \bf{78.1}           & 12.80  
    & \scriptsize[224,224] [102,99] [43,41] [16,17] [4,4]       
    \\ \cmidrule{1-11}\morecmidrules\cmidrule{1-11}
    
    \multicolumn{1}{c}{\multirow{4}{*}{\rotatebox{90}{MBN-V2}}~~}
    & Human      
    & 77.6            & 0.33        
    & \scriptsize[224,224] [112,112] [56,56] [28,28] [14,14] [7,7]   
    & 73.8           & 0.09        
    & \scriptsize[32,32] [16,16] [8,8] [4,4]
    & 71.7           & 0.31 
    & \scriptsize[224,224] [112,112] [56,56] [28,28] [14,14] [7,7]        
    \\ \cline{2-11} 
    \\ [-0.9em]
    & DynOPool-S
    & 78.7            & 0.34        
    & \scriptsize[224,224] [98,119] [39,42] [36,18] [21,9] [12,4]
    & 74.0            & 0.08        
    & \scriptsize[32,32] [13,13] [6,6] [4,4]
    & 72.1           & 0.49 
    & \scriptsize[224,224] [111, 111] [55,50] [32,27] [20,16] [9,7]        
    \\
    \\ [-0.9em]
    & DynOPool-B
    & \bf{82.6}            & 2.35       
    & \scriptsize[224,224] [181,150] [132,174] [87,80] [51,36] [22,13] 
    & \bf{76.2}         & 0.21        
    & \scriptsize[32,32] [22,21],[12,12] [7,7]
    & \bf{73.8}         & 1.16 
    & \scriptsize[224,224] [181,171] [95,93] [53,53] [31,29] [10,10]          
    \\ \bottomrule
    \end{tabular}
    }
    \end{table*}

\subsection{Versatility of DynOPool}
\label{sec:dynopool_networks}

Due to its model-agnostic property, DynOPool can replace all kinds of resizing operators in any given network.
To analyze the superiority of the optimized scale factor $\bm{r}$ to the predetermined methods relying on hyperparameters, we replace all types of resizing operators in the baseline network with DynOPool except for the last global average pooling layer; pooling operations (\eg max-pooling) are replaced by DynOPool and strided convolutions are replaced by the combinations of a vanilla convolution (with stride 1) and DynOPool. For a detailed description of each model, please refer to the supplementary document.

Unlike other methods that require to select either downsampling or upsampling in advance and depend on the pre-defined pooling ratios, DynOPool learns to resize feature maps without the constraint for the scale factor and the pooling ratio.
In practice, the upsampling process of DynOPool is the same as the downsampling.
A tricky thing in upsampling is that it can use the features of the same set of pixels to calculate the features of different query points.
However, it does not incur any issue because the distances to the pixels from each query point are different and the features for each query point are different.


\section{Experiments}
\label{sec:experiments}
This section summarizes the experimental results with DynOPool on various types of networks and datasets.
For the classification task, we use three datasets and three types of networks for evaluation. 
We compare our model with human-designed models and Shape Adaptor~\cite{liu2020shape} in terms of accuracy and GMACs, and present that dynamic resizing layers boost performance with almost no extra cost.
Furthermore, we apply our module to EfficientNet~\cite{tan2019efficientnet} to show compatibility to the NAS algorithms, and conduct an additional experiment on PascalVOC~\cite{pascalvoc} to prove applicability to the semantic segmentation task.

\subsection{Experiment Setup}

\paragraph{Models}
We mainly apply DynOPool to three baselines: VGG-16~\cite{simonyan2015very}, ResNet-50~\cite{he2016deep}, and MobileNetV2~\cite{sandler2018mobilenetv2}.
We also use EfficientNet-B0~\cite{tan2019efficientnet} to check compatibility with NAS.
DynOPool is adopted to the downscaling module of each model and keeps the rest of the structure the same as the human-designed architecture.
It is worth noting that there is no increase in the number of parameters of the models with DynOPools except for the scale parameter $\bm{\alpha}$. 

\paragraph{Datasets}
We conduct the experiment on three datasets including FGVC-Aircraft~\cite{maji2013fine}, CIFAR-100~\cite{krizhevsky2009learning} and ImageNet~\cite{russakovsky2015imagenet}.
Unlike CIFAR-100 and ImageNet that contain diverse general objects, Aircraft is a fine-grained dataset for aircraft classification.
CIFAR-100 is a dataset with small (32 $\times$ 32) images while the size of the images in  Aircraft and ImageNet are large (224 $\times$ 224).
The experiment setting is to verify that DynOPool performs well regardless of image sizes or data characteristics.

\paragraph{Implementation details}
For optimization, we employ the same hyperparameters as Shape Adaptor except for the number of epochs.
According to our experience, DynOPool requires more epochs than Shape Adaptor for training to allow both the scale factor and weights to converge sufficiently in response to the dynamic model structure changes.
Especially, CIFAR-100 and Aircraft, which have relatively small datasets, are greatly affected by the epoch.
Accordingly, we increase the epoch from 200 to 250 for models with DynOPool on both datasets.

The learning rate for the scale parameter $\bm{\alpha}$ is lower than that of the model parameter similar to other dynamic networks~\cite{dai2017deformable,liu2020shape} since the scale parameter affects the entire model even with its slight changes.
To prevent the feature map size from reducing to 1 during training, we bound the output feature map size by $H_{\text{out}} = \round{\max(H_{\text{in}} \cdot r_h, 1.5)}$, which ensures that the size of a feature map is at least 2 in each dimension while allowing a model to backpropagate gradients through the feature map in any dimension smaller than 2.
For other hyperparameters and experimental settings, we list details in the supplementary document.

\begin{figure*}[t]
	\centering
	\footnotesize
	\setlength{\tabcolsep}{15pt}
	\begin{tabular}{cccc}		
		\includegraphics[height=4.6cm]{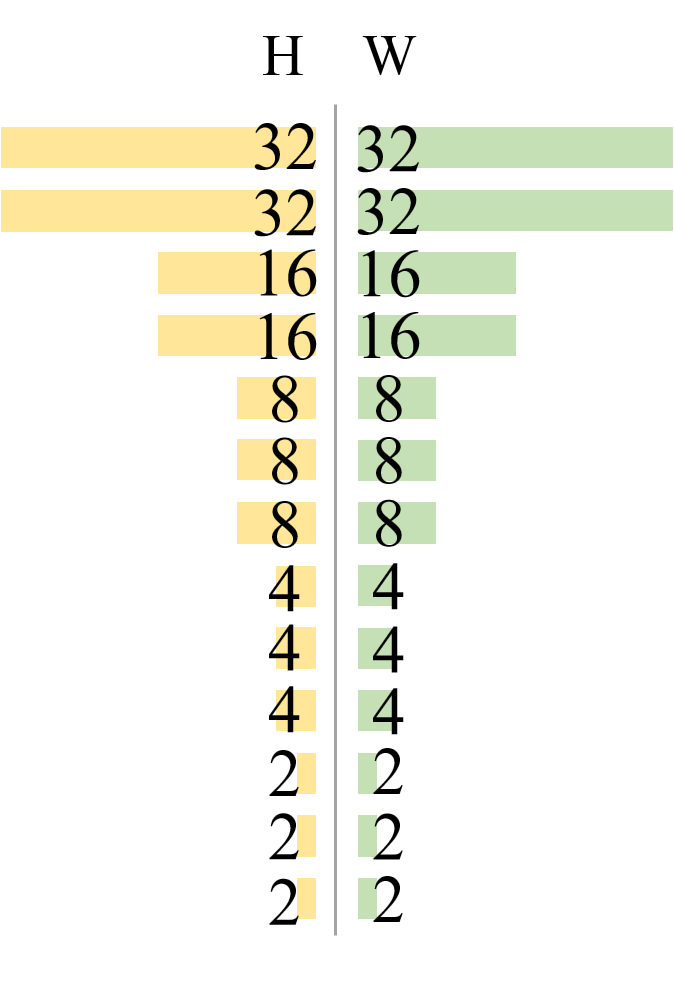} &
		\includegraphics[height=4.6cm]{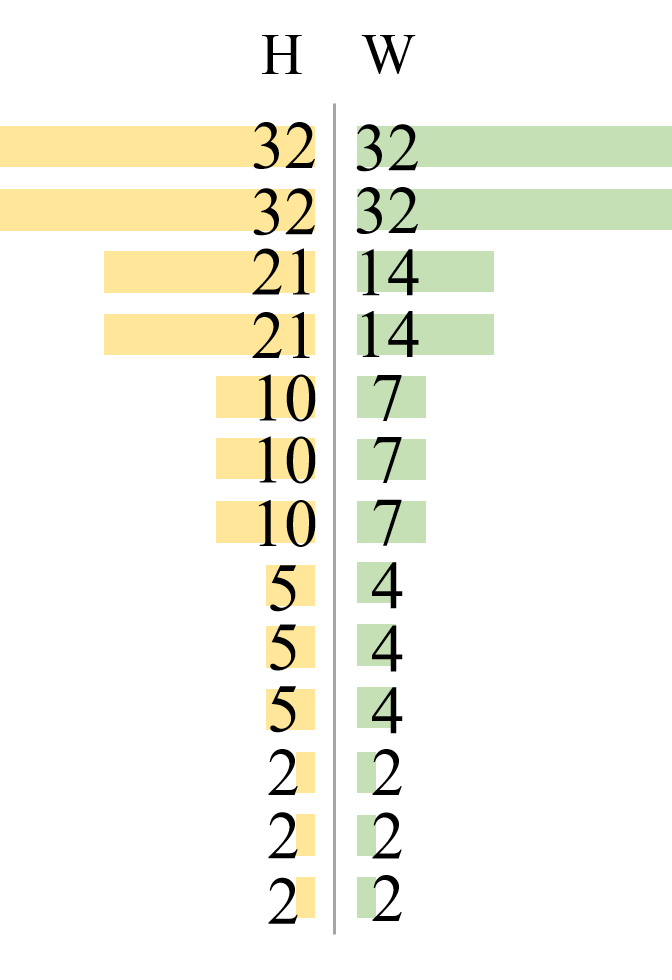} &	
		\includegraphics[height=4.6cm]{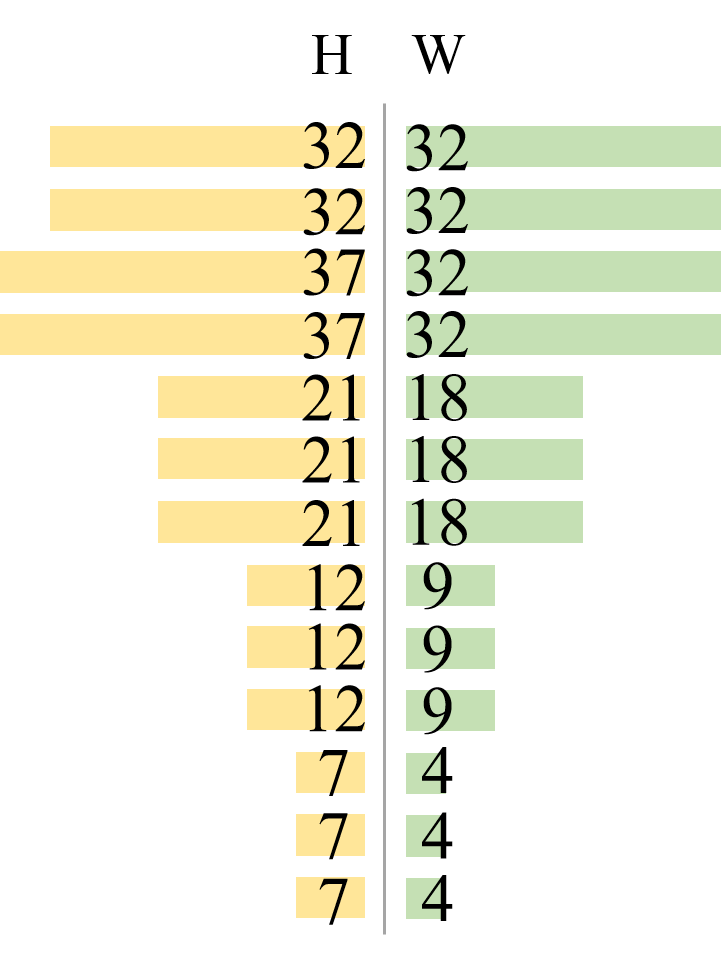} &
		\includegraphics[height=4.6cm]{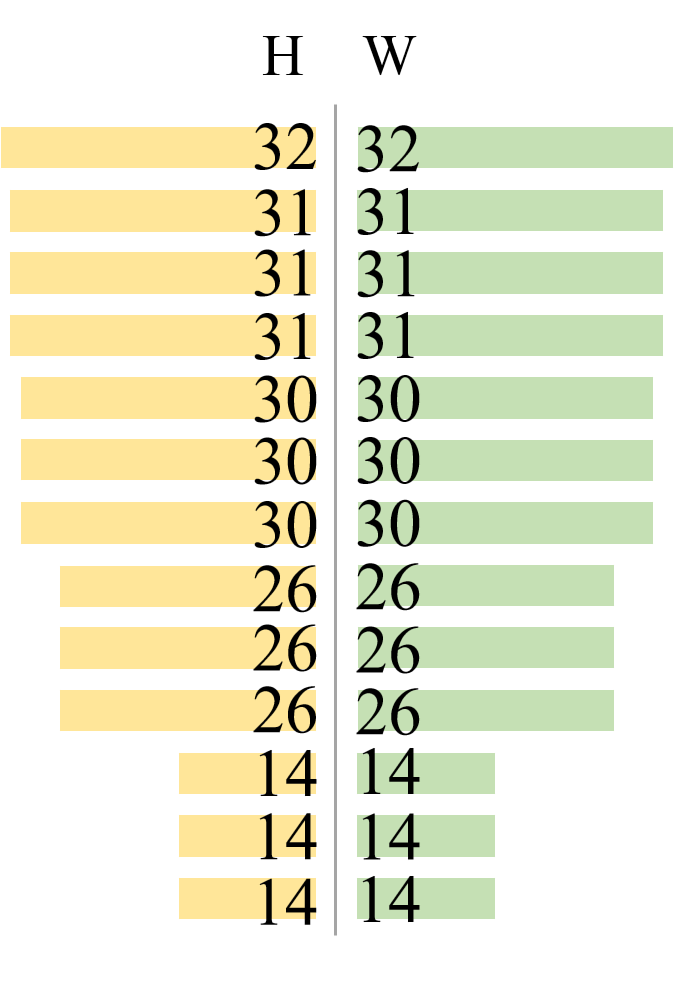} \\
		\shortstack{0.31G, 75.4\% \\ (a) Human-designed } 
		& \shortstack{0.36G, 75.5\% \\ (b) DynOPool-S} 
		&  \shortstack{1.71G, 79.8\% \\ (c) DynOPool-B} 
		& \shortstack{5.21G, 79.2\% \\ (d) Shape Adaptor}		
		\end{tabular}
		\caption{Visualization of trained models with DynOPool and Shape Adaptor from the human designed VGG-16 on the CIFAR-100 dataset. We visualize the sizes and shapes of intermediate feature maps in each model with GMACs and accuracy. By learning the optimal scale parameter $\bm{\alpha}$ for the dataset, DynOPool illustrates competitive performance compared to the human-designed model and Shape Adaptor.}
	\label{fig:vofs}
\end{figure*}	

\subsection{Comparison with Human-Designed Model} 
We discuss the performance and the characteristics of the proposed approaches in comparison with the human-designed models.

\subsubsection{Main results}

Table~\ref{tab:main} presents the performance of DynOPool in terms of GMACs and accuracy. 
We compare the human-designed model with two variants of our model with DynOPool: 1) a model with a small computational cost similar to that of the human-designed model (DynOPool-S) and 2) a model learned mainly for accuracy  (DynOPool-B).

DynOPool-S improves accuracy significantly with almost the same or fewer GMACs as the human-designed model in most cases, and DynOPool-B outperforms the human-designed model in all settings.
Note that we greatly improve the performance by changing the size and shape of feature maps with little increase in the number of parameters.
To achieve this goal with NAS, it would take at least a few dozen GPU days since the search space is huge due to a large number of resizing layers and the consideration of information asymmetry.
On the contrary, DynOPool solves the above problem successfully and identifies an optimized network without an exhaustive search process.

On FGVC-Aircraft, as presented in Table~\ref{tab:main}, trained networks have many non-square feature maps with the receptive fields of the reciprocal shapes and achieve the largest performance improvements among all the tested datasets.
Since the images in the fine-grained dataset share relatively many patterns in common than in general images, it may be critical to find the optimal shape of the receptive field to achieve better accuracy. 
It is interesting that DynOPool-S models have wide feature maps in the early layers but end up with tall feature maps in the deeper layers.
This fact implies that the proposed dynamic resizing modules concentrate on the information in the horizontal direction in analyzing local patterns, which forces the information in the vertical direction to become more important in identifying semantic structures in images.
As a result, it turns out to achieve superior performance with less computation than the human-designed models relying on the feature maps with the standard sizes and shapes.

Table~\ref{tab:main} illustrates another interesting results about the feature map shapes of the networks trained on CIFAR-100 and ImageNet, which contain more general object categories in images than FGVC-Aircraft.
The feature maps are optimized for vertical shapes, \ie, $H > W$, in almost all settings, which also aligns with the result from the previous work~\cite{riad2022learning}.
This implies that the amount of information in the ImageNet and CIFAR-100 datasets have is asymmetric in the spatial dimensions and we can extract more information by observing the details in the vertical direction than in the horizontal direction.

Furthermore, we visualize the feature map sizes of the human-designed model, DynOPool-S/B, and Shape Adaptor in Figure~\ref{fig:vofs}.
As shown in Figure~\ref{fig:vofs}(b) and (c), DynOPool-S/B learn to utilize non-square feature maps and exhibit the data-driven model selection capability. 
In particular, DynOPool-B even increases the feature map size after the first pooling layer, which leads to substantial accuracy gain by 4.4\%p compared to the human-designed model..
The result of DynOPool-B shows that the full use of local information in the front layers is sometimes helpful while enlarging the receptive field size later to reduce the sizes of the corresponding feature maps.

\begin{figure}[t]
    \centering
    \includegraphics[width=0.9\linewidth]{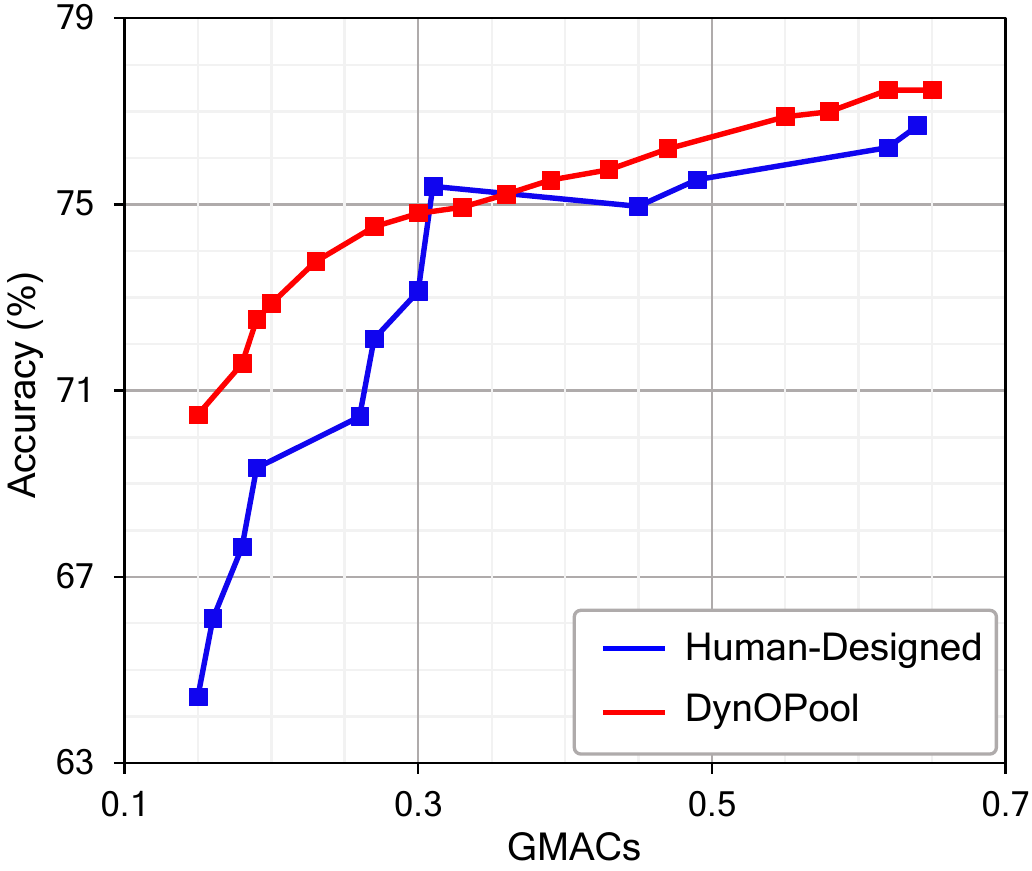}
    \caption{GMACs-Accuracy tradeoffs between human-designed VGG-16 and VGG-16 with DynOPool on CIFAR-100. The models with DynOPool are trained with different values of $\lambda$ while human-designed models are trained by varying input resolutions.}
    \label{fig:gac}
\end{figure}

\subsubsection{Trade-off between accuracy and GMACs}

Figure~\ref{fig:gac} illustrates the GMACs-accuracy tradeoffs between our model with DynOPool and human-designed model with VGG-16 on CIFAR-100.
We adjust the input image size to obtain the accuracies of the human-designed model, VGG-16 with respect to different computational cost in terms of GMACs. 
This is motivated by the strategy of several NAS algorithms that include the input size in the search space~\cite{liu2018darts, tan2019efficientnet}.
For DynOPool, we control GMACs by varying the coefficient for the GMACs loss $\lambda$ in \eqref{eq:loss_total}.

DynOPool shows superior trade-off between accuracy and GMACs compared to the human-designed model in almost all cases, especially when the models are compressed significantly.
This is because, by using our approach, the model structure is optimized dynamically and effectively for the target GMACs.
In the case of the human-designed model, the performance is optimized with a good trade-off when the input image size is exactly $32 \times 32$ (0.31 GMACs). 
We believe that this is because the CIFAR-100 dataset has been tested extensively for years using its original image size and most of the human-designed models are optimized best for the input size.
Also, the human-designed models may not be effective to handle non-conventional input image sizes other than numbers to the power of 2 due to the potential errors given by extra paddings.

\subsection{Comparison with Shape Adaptor}


\begin{table}[t]
    \centering
    \small
    \caption{Comparison between DynOPool and Shape Adaptor on the CIFAR-100 dataset. DynOPool consistently outperforms Shape Adaptor with lower computational costs.}
    \label{tab:sa}
    \scalebox{0.9}{
    \renewcommand{\arraystretch}{0.95}
    \setlength\tabcolsep{5pt}
    \begin{tabular}{cccc}
    \hline
    
    \toprule
    \\ [-1.5em]
    \multicolumn{1}{c}{\multirow{2}{*}{\textbf{Backbone}}}
    & \multicolumn{1}{c}{\multirow{2}{*}{\textbf{Model}}}
    & \multicolumn{1}{c}{\multirow{2}{*}{Acc.}}
    & \multicolumn{1}{c}{\multirow{2}{*}{GMACs}}
    \\ 
    \multicolumn{1}{c}{} & \multicolumn{1}{c}{} 
    & \multicolumn{1}{c}{} & \multicolumn{1}{c}{}
    \\ [-1.5em]
    \\ \cmidrule{1-4}\morecmidrules\cmidrule{1-4}
    
    \multicolumn{1}{c}{\multirow{2}{*}{VGG-16}}
    & Shape Adaptor      
    & 79.2            & 5.21            
    \\
    \\ [-1.0em]
    & DynOPool (ours)
    & $\textbf{79.8}$            & \textbf{1.71}         
    \\ [-0.9em]
    \\ \hline
    \\ [-0.9em]
    
    \multicolumn{1}{c}{\multirow{2}{*}{\begin{tabular}[c]{@{}c@{}} {} \vspace{-3mm}\\ ResNet-50\end{tabular}}}
    & Shape Adaptor   
    & 80.3            & 4.93        
    \\
    \\ [-1.0em]
    & DynOPool (ours)
    & \textbf{80.6}            & \textbf{1.73}     
    \\ [-0.9em]
    \\ \hline
    \\ [-0.9em]
    
    \multicolumn{1}{c}{\multirow{2}{*}{\begin{tabular}[c]{@{}c@{}} {} \vspace{-2.5mm}\\ MobileNetV2\end{tabular}}}
    & Shape Adaptor        
    & 75.7            & 0.92            
    \\
    \\ [-1.0em]
    & DynOPool (ours)  
    & \textbf{76.2}            & \textbf{0.21}       
    \\ [-1.0em]
    \\ \bottomrule
    \end{tabular}
    }
    \end{table}

Table~\ref{tab:sa} compares the accuracy and the GMACs between DynOPool and Shape Adaptor~\cite{liu2020shape}.
Although both algorithms aim to find the optimal feature map sizes by introducing learnable resizing modules, DynOPool outperforms Shape Adaptor in terms of both accuracy and efficiency.
We demonstrate the feature map sizes of Shape Adaptor in Figure~\ref{fig:vofs}(d) together with the accuracy and the computational complexity.

We believe that the following trait of our method drives the difference.
Shape Adaptor determines the output feature map size by a linear interpolation of two pre-defined candidate size scales.
This strategy results in large approximation errors by forcibly considering potentially irrelevant features for aggregations under the predicted scale factor.
On the contrary, DynOPool adjusts the feature map size naturally using a single scale factor $\bm{r}$, which is reparametrized by $\bm{\alpha}$ for stable optimization.
More detailed comparisons between the two approaches are discussed in the supplementary document.


\begin{table}[t]
\centering
\small
\caption{Performance of DynOPool with EfficientNet-B0 on the ImageNet dataset.}
\label{tab:eff}
\scalebox{0.9}{
\setlength\tabcolsep{4pt}
\begin{tabular}{cccc}
\hline \toprule
\textbf{Backbone} & \textbf{Model}    & Acc.          & GMACs        \\ \hline\hline \\[-0.9em]
EfficientNet-B0 & \multirow{2}{*}{Human}               & 71.8         & 0.42          \\ \\[-0.9em]
EfficientNet-B1  &                                                    & 72.8        & 0.75 \\ \hline  \\[-0.9em]
EfficientNet-B0  & DynOPool (ours)                                   & 72.3        & 0.58        \\ \bottomrule
\end{tabular}
}
\end{table}

\subsection{Compatibility with NAS algorithms}
While NAS is a more general concept than DynOPool, the feature map size is not typically considered in the search space in NAS and the architecture can be optimized by NAS jointly with DynOPool.
We adopt DynOPool for the optimization of EfficientNet~\cite{tan2019efficientnet}, which is one of the state-of-art architectures identified by NAS.
As seen in Table~\ref{tab:eff}, EfficientNet-B0 with DynOPool shows competitive performance in terms of accuracy and GMACs compared to both EfficientNet-B0 and EfficientNet-B1.

Although the benefit of DynOPool is not impressive in this result, the combination of NAS and DynOPool are formulated as a differentiable optimization task even in the feature map scale dimension; it has potential to lead to higher accuracy with less computational cost. 
Note that, while the architecture of EfficientNet-B1 is identified from a combinatorial search space in 1) width, 2) depth, and 3) resolution dimensions, we can find competitive models with the optimized feature map sizes at a substantially reduced search time using DynOPool.


\begin{table}[t]
\centering
\small
\caption{
Semantic segmentation results of HRNet-W48 on PascalVOC.
DynOPool compresses the human-designed model up to 16\% with slight improvement of mIoU.
}
\label{tab:seg}
\scalebox{0.9}{
\setlength\tabcolsep{2.5pt} 
\begin{tabular}{cccl}
\hline
\toprule
\textbf{Model} & mIoU & GMACs & \hspace{11mm}{Feature map sizes}
\\ \cmidrule{1-4}\morecmidrules\cmidrule{1-4}
Human      & 76.2            & 82.55            & \scriptsize[240,240] [120,120] [60,60] [30,30] [15,15] \\ \\[-0.9em]
DynOPool (ours) & \bf{76.4}           & \bf{69.39}        & \scriptsize[367,349] [134,130] [52,50] [22,21] [10,9]\\
\bottomrule
\end{tabular}
}
\end{table}

\subsection{Semantic Segmentation Results}
To further verify the effectiveness of DynOPool, we conduct additional experiments on semantic segmentation.
The semantic segmentation task involves various objects and stuff in a scene with various scales, identifying the optimal receptive field corresponding to each object is critical to improve the final accuracy.
To get semantically richer and spatially more precise representations, multi-scale representation learning is is the prevalent approach in semantic segmentation models~\cite{cai2016unified, zhao2017pyramid, chen2018encoder, WangSCJDZLMTWLX19}.
For example, HRNet~\cite{WangSCJDZLMTWLX19} maintains high-resolution representations throughout the whole process and connects the high-to-low resolution convolution streams in parallel. 

To evaluate the performance of DynOPool in semantic segmentation, we employ HRNet-W48, a variant of HRNet, as our backbone model, and replace the strided convolutions in the model by a combination of DynOPool and a vanilla convolution (with stride 1).
We train the models on the PascalVOC~\cite{pascalvoc} dataset to check if there exists further room for improvement.
As seen in Table~\ref{tab:seg}, DynOPool successfully compresses the human-designed model up to 16\% with a slight improvement of the mIoU.
Interestingly, our model enlarges the resolution of the convolution stem and the upper branch of the parallel convolution stream, and consistently reduces the resolution of the remaining three branches of parallel convolution streams.
This highlights the importance of maintaining the feature maps with data-driven feature map sizes to improve performance with less computational burden.
We present detailed experimental settings in the supplementary document.

\section{Conclusion and Future Works}
\label{sec:conclusion}
\subsection{Conclusion}
We presented a Dynamically Optimized Pooling, referred to as DynOPool, which facilitates finding an optimized sizes and shapes of receptive fields and feature maps.
DynOPool identifies the optimal size and shape of feature maps without relying on human inductive bias or exhaustive architecture search.
Our module achieved superior performance with various recognition models on multiple datasets, and showed desirable trade-offs between accuracy and computational cost, compared to the human-designed model and the previous work.
We also showed that DynoPool is compatible with the recent NAS algorithms and naturally applicable to semantic segmentation model.
We hope that our module allows the vision community to optimize deep neural networks more effectively.

\subsection{Future Works}
Although we focus on the two-dimensional tasks in this work, our module could be extended to higher dimensional scaling modules.
For example, in an action recognition task, we can also employ DynOPool to capture temporal relation from a dataset by adjusting the number of frames required for temporal pooling.

Furthermore, similar to our findings, in cognitive science, it has been well-known for decades that the human visual system perceives vertical lines to be slightly longer than horizontal ones~\cite{finger1947illustration, kunnapas1955analysis,robinson2013psychology} and judge 
the symmetry based more on the horizontal symmetry than the vertical counterpart~\cite{fisher1987goldmeier, rock1963experimental}.
In other words, our visual system has been adapted to be more sensitive to vertical information changes.
Despite the long history, the exact cause has not yet been identified and is still under discussion~\cite{berry2002cross, mamassian2010simple}.
It would be worthwhile to investigate the connection between the findings from our work and the observations in cognitive science, which makes a synergy to understand the asymmetric behavior of computer vision and human vision systems and bridges a missing link between two research fields.


\paragraph{Acknowledgement}

Dong-Hwan Jang is grateful for financial support from Hyundai Motor Chung Mong-Koo Foundation.
This research was supported in part by Samsung Advanced Institute of Technology and by the Bio \& Medical Technology Development Program of the National Research Foundation (NRF) [No. 2021M3A9E4080782] and Institute of Information \& communications Technology Planning \& Evaluation (IITP) grants [No.2021-0-01343, Artificial Intelligence Graduate School Program (Seoul National University) ; No.2021-0-02068, Artificial Intelligence Innovation Hub] funded by the Korea government (MSIT).

\clearpage

{\small
\bibliographystyle{ieee_fullname}
\bibliography{egbib}
}


\onecolumn

\setcounter{section}{0}
\setcounter{table}{0}
\setcounter{figure}{0}
\setcounter{equation}{0}
\renewcommand\thesection{\Alph{section}}
\renewcommand\thetable{\Alph{table}}
\renewcommand\thefigure{\Alph{figure}}
\renewcommand\theequation{\Alph{equation}}


\section{Implementation Details}

\subsection{DynOPool Networks}
For a fair comparison, we use the same model architecture and augmentation strategies as Shape Adaptor~\cite{liu2020shape}.
Following the prior work, only a single fully-connected layer is used at the end of CNNs for both the human-designed and DynOPool-equipped models to concentrate on the effect of feature shapes.

For classification, we replace the resizing modules (the pooling and strided convolution layers) as described below.
\begin{equation}
    \text{Pooling}_s\rightarrow\text{DynOPool}_s,
\end{equation}
\begin{equation}
    \text{ReLU}\circ\text{Conv}_s\rightarrow\text{DynOPool}_s\circ\text{ReLU}\circ\text{Conv}_1.
    \label{eq:strided_conv}
\end{equation}
Here, a subscription $s$ represents the stride for each operation.
In ResNet-50~\cite{he2016deep} for ImageNet~\cite{russakovsky2015imagenet}, the pooling layer comes right after the strided convolution layer in the first two blocks.
In such a case, DynOPool is applied twice in a row (refer to (\ref{eq:strided_conv})) and we use one DynOPool having 0.25 as the initial scaling factor to prevent the training instability.
Moreover, if there are more than two branches with the same resolution as in ResNet, the same scale parameter $\bm{\alpha}$ is shared across the branches to ensure consistency. 

For semantic segmentation, we plug-in DynOPool into HRNet-W48~\cite{WangSCJDZLMTWLX19} as follows:
\begin{equation}
    \text{Conv}_s\rightarrow\text{Conv}_1\circ\text{DynOPool}_s.
\end{equation}
In HRNet-W48, each convolution stream contains the strided convolution layers for downsampling and the bilinear interpolation layers for upsampling.
We replace the strided convolution layers with DynOPool + vanilla convolution (of stride 1) and maintain the bilinear interpolation layers.
Since the resolution of the feature maps should be the same for each level of the convolution stream in HRNet-W48, we share the same DynOPool module instance within each convolution stream.

\subsection{Experimental Setup}
We list all ingredients and hyperparameters for our models in Table~\ref{tab:flops_coeff} and~\ref{tab:hparam}.
Initially, we set the scale factor $\bm{r}$ of DynOPool to 0.5, which is the same as the human-designed model and adjust the model size through $\lambda$.
However, in cases of ResNet-50 on CIFAR-100 and Aircraft, we observe that model size does not increase even when $\lambda=0$. 
In order to measure the performance of the optimal shape by model size, we increased the initial scale factor of the first layer in these cases.
We list up $\lambda$s and initial scale factors of the first resizing layer for all settings in Table~\ref{tab:flops_coeff}.

For the experiments in Section 3 of the main paper (CIFAR-stretch/tile/large), we use the same hyperparameters as VGG-16~\cite{simonyan2015very} on CIFAR-100~\cite{krizhevsky2009learning}. 
To regularize the model sizes, $\lambda$s are set to 2.33e-4/7.21e-5/5.03e-5, respectively.
For EfficientNet-B0 with DynOPool, we use the same hyperparameters as VGG-16 on ImageNet. Also, $\lambda$ is set to 2.50e-5.

The semantic segmentation model is trained for 100 epochs with a batch size of 16 on a single GPU.
The initial learning rate is set to 1.6e-2 for the weights and 8e-3 for the $\bm{\alpha}$ with the weight decay of 1e-4. Also, $\lambda$ is set to 7.00e-5.
As an optimizer, we use SGD with a momentum 0.9 and a polynomial learning rate policy with the power of 0.9.

We implement all experiments with PyTorch~\cite{NEURIPS2019_9015}, and use the Automatic Mixed Precision package for the ImageNet experiments.

\begin{table}[b]
    \centering
    \caption{Coefficient for GMACs and learning rate for shape parameter $\bm{\alpha}$ for each model.}
    \label{tab:flops_coeff}
    \scalebox{0.9}{
    \setlength\tabcolsep{3.5pt} 
    \begin{tabular}{@{}lcccccccccc@{}}
    \hline
    \toprule
    \\ [-0.9em]
    \multicolumn{2}{c}{\multirow{2}{*}{}}
    & \multicolumn{3}{c}{\textbf{FGVC-Aircraft}} 
    & \multicolumn{3}{c}{\textbf{CIFAR-100}} 
    & \multicolumn{3}{c}{\textbf{ImageNet}} 
    \\ \cmidrule(l){3-11}
    \\ [-0.9em]
    
    \multicolumn{2}{c}{}
    & VGG-16     & ResNet-50     & MBN-V2
    & VGG-16     & ResNet-50     & MBN-V2
    & VGG-16     & ResNet-50     & MBN-V2
    \\ \cmidrule{1-11}\morecmidrules\cmidrule{1-11}
    \\ [-0.9em]
    
    \multirow{3}{*}{DynOPool-S}
    & $\lambda$ 
    & 3.07e-5 & 3.99e-5 & 3.00e-5 
    & 2.20e-4 & 0       & 5.00e-5
    & 1.00e-4 & 8.00e-5 & 4.00e-5
    \\ \cmidrule(l){2-11}
    & \begin{tabular}[c]{@{}c@{}}Init. Scale\\ (first layer)\end{tabular}
    & 0.5 & 0.5 & 0.5
    & 0.5 & 0.6 & 0.5
    & 0.5 & 0.5 & 0.5
    \\ \midrule
    \\ [-0.9em]
    
    \multirow{3}{*}{DynOPool-B}
    & $\lambda$
    & 6.51e-6 & 0       & 0 
    & 1.59e-5 & 0       & 1.00e-7
    & 7.00e-5 & 2.00e-5 & 1.00e-5
    \\ \cmidrule(l){2-11}
    & \begin{tabular}[c]{@{}c@{}}Init. Scale\\ (first layer)\end{tabular}
    & 0.5 & 0.7 & 0.5
    & 0.5 & 0.8 & 0.5
    & 0.5 & 0.5 & 0.5
    \\ \bottomrule
    \\ [-0.9em]
    \end{tabular}
    }
    \end{table}

\begin{table}[t]
    \centering
    \caption{Hyperparameter list for the experiments.}
    \label{tab:hparam}
    \scalebox{0.9}{
    \setlength\tabcolsep{3.5pt}
    \begin{tabular}{@{}lccccccccc@{}}
    \hline
    \toprule
    \\ [-0.9em]
    \multirow{2}{*}{} 
    & \multicolumn{3}{c}{\textbf{FGVC-Aircraft}}
    & \multicolumn{3}{c}{\textbf{CIFAR-100}} 
    & \multicolumn{3}{c}{\textbf{ImageNet}} 
    \\ \cmidrule(l){2-10}
    \\ [-0.9em]
    
    & VGG-16     & ResNet-50    & MBN-V2     
    & VGG-16     & ResNet-50    & MBN-V2
    & VGG-16     & ResNet-50    & MBN-V2
    \\ \cmidrule{1-10}\morecmidrules\cmidrule{1-10}
    \\ [-0.9em]
    Learning Rate     
    & \multicolumn{3}{c}{1e-2}
    & \multicolumn{3}{c}{1e-1}
    & 1e-1     & 1e-1     & 5e-2            
    \\ \midrule
    \\ [-0.9em]
    Learning Rate ($\bm{\alpha}$) 
    & \multicolumn{3}{c}{1e-2}
    & \multicolumn{3}{c}{1e-2}
    & \multicolumn{3}{c}{5e-3}          
    \\ \midrule
    \\ [-0.9em]
    Optimizer         
    & \multicolumn{9}{c}{SGD with 0.9 momentum}
    \\ \midrule
    \\ [-0.9em]
    Scheduler         
    & \multicolumn{9}{c}{Cosine Annealing}
    \\ \midrule
    \\ [-0.9em]
    Weight Decay      
    & \ 5e-4 & \ 5e-4 & \ 4e-5
    & \ 5e-4 & \ 5e-4 & \ 4e-5
    & \ 5e-4 & \ 5e-4 & \ 4e-5
    \\ \midrule
    \\ [-0.9em]
    Batch Size        
    & \multicolumn{3}{c}{8}
    & \multicolumn{3}{c}{128}                      
    & \multicolumn{3}{c}{64 (per GPU) for 4 GPUs}   
    \\ \midrule
    \\ [-0.9em]
    Epochs            
    & \multicolumn{3}{c}{250}
    & \multicolumn{3}{c}{250}                      
    & \multicolumn{3}{c}{120}                       
    \\ \bottomrule
    \\ [-0.9em]
    \end{tabular}
    }
    \end{table}
    
\section{Comparison with Shape Adaptor}

\begin{figure*}[h]
\centering
  \includegraphics[width=.7\linewidth]{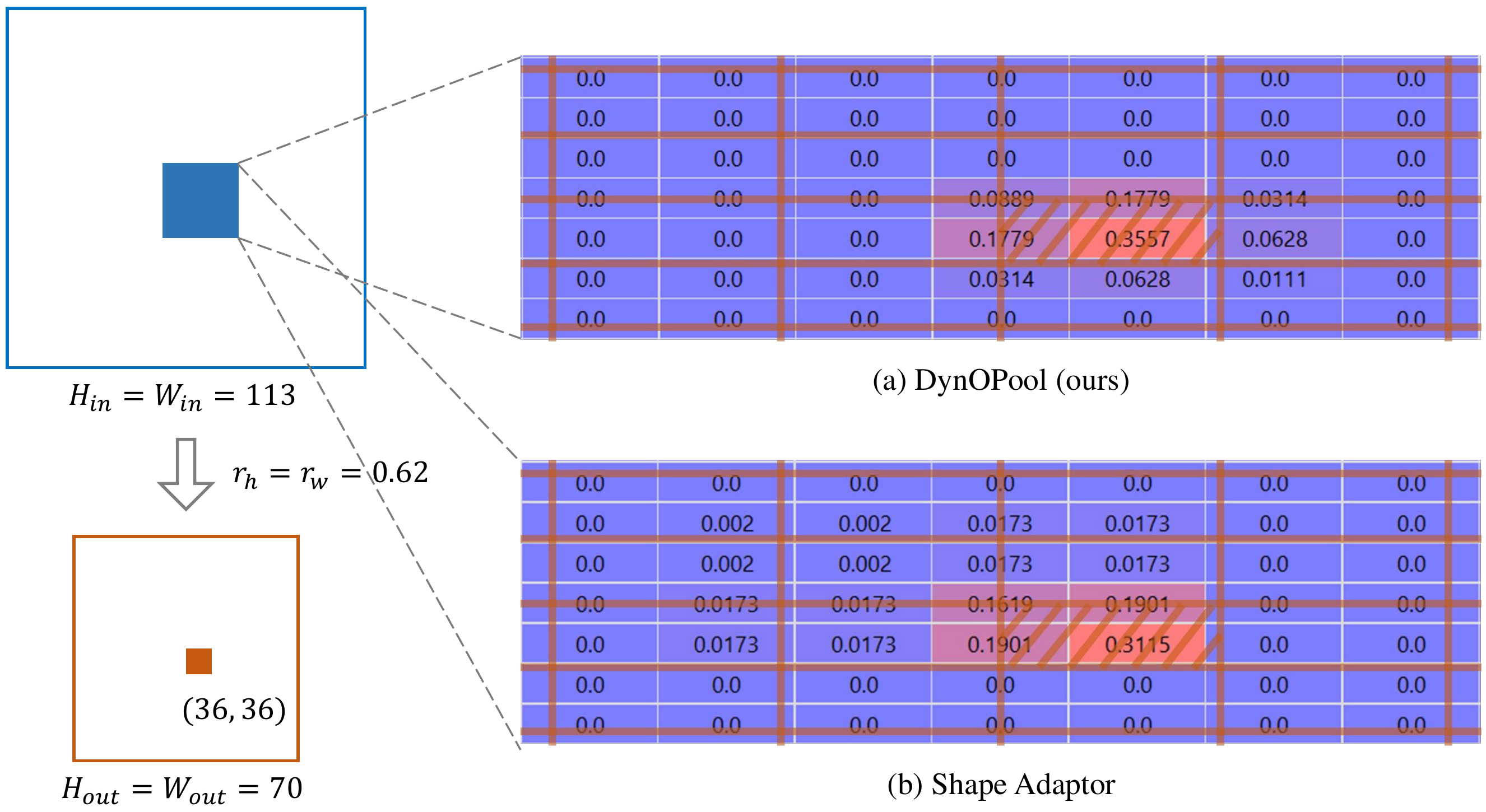}
\caption{\textbf{Visualization of feature aggregation ranges} (a) DynOPool (ours) and (b) Shape Adaptor.
The brown grid represents the grid of output features overlaid on the input feature.
The numbers written in each bin of the input feature indicate the contribution of each feature to compose a shaded output bin.
Since Shape Adaptor finds the optimal pooling ratio by linear interpolation between two branches with different pre-defined pooling ratios, its aggregation range is highly quantized and unrelated features are involved.}
\label{fig:comparison}
\end{figure*}

In Figure~\ref{fig:comparison}, we visualize the feature aggregation range for both Shape Adaptor~\cite{liu2020shape} and ours.
Shape Adaptor uses linear interpolation between the features with the pre-defined rescaling ratios to find the optimal pooling ratio. However, since the resulted ratio lies between two pre-defined ratios, its receptive field gets bigger than it needs. 

Furthermore, since Shape Adaptor uses pooling before the aggregation, the overall performance is closely affected by the pre-defined pooling ratio.
Therefore, setting a small pooling ratio smaller than 0.5, to broaden the search space of the pooling ratio, could potentially harm the overall performance.
Since we use less quantization for feature selection, each bin of the output feature can aggregate more closely related features from the input.

\section{Resource Efficiency}

\begin{table}[h]
    \centering
    \small
    \caption{The relative computational cost of DynOPool compared to the entire model. DynOPool brings negligible computational overheads compared to other operators in the networks.}
    \label{tab:cost}
    \scalebox{0.9}{
    \begin{tabular}{@{}cccc@{}}
    \toprule
    \textbf{Dataset}    & \textbf{FGVC-Aircraft} & \textbf{CIFAR-100} & \multicolumn{1}{l}{\textbf{ImageNet}} \\ \cmidrule{1-4}\morecmidrules\cmidrule{1-4}
    VGG-16      & 0.07 \% & 0.07 \%              & 0.07 \%                                  \\
    ResNet-50   & 0.26 \% & 0.13 \%               & 0.26 \%                                  \\
    MobileNetV2 & 3.88 \% & 1.00 \%               & 3.86 \%                                  \\ \bottomrule
    \end{tabular}
    }
    \end{table}
In Table~\ref{tab:cost}, we calculate the percentage of DynOPool's GMACs compared to the overall GMACS of the entire models.
The amount of computational cost brought by DynOPool is only a fraction of the total cost.
For VGG-16 and ResNet-50, our modules' costs are less than 1\% of the entire models' costs.
Even in MobileNetV2, where the convolution operation is relatively light, the percentages do not exceed 5\%.

The reason for the low computational costs of our modules can be two-fold. First, the convolutional operation is much more expensive than the operation of bilinear interpolation.
The amount of computation for convolution is proportional to $C_{\text{in}} \cdot C_{\text{out}} \cdot H_{\text{out}} \cdot W_{\text{out}}$, while the cost of bilinear interpolation is proportional only to the size of the output, $C_{\text{out}} \cdot H_{\text{out}} \cdot W_{\text{out}}$, where $C_{\text{in}}$ and $C_{\text{out}}$ are the number of channels of the input and output, respectively.
Second, the number of pooling layers is much smaller than the number of convolution layers. For example, there are only four pooling layers in VGG-16.

\section{Limitations}
Since the model size is indirectly adjusted through $\lambda$, there is a limitation in finding a model of the exact target GMACs.
Also, we could only estimate the distribution of information in each dataset through the obtained feature shapes.
However, it would be great if future research could provide a way to numerically analyze this.

\end{document}